\documentclass[letterpaper, 10pt, journal, twoside]{IEEEtran}

\usepackage{amsmath,amssymb,amsfonts}
\usepackage{algorithmic}
\usepackage{algorithm}
\usepackage{adjustbox}
\usepackage{tabularx}
\usepackage{bm}
\usepackage{array}
\usepackage[caption=false,font=footnotesize]{subfig}
\usepackage{textcomp}
\usepackage{stfloats}
\usepackage{url}
\usepackage{verbatim}
\usepackage{graphicx}
\usepackage{cite}
\usepackage{rotating}
\usepackage{booktabs}
\usepackage{xcolor}
\usepackage{pgf}
\usepackage{tikz}
\usepackage{orcidlink}
\definecolor{spec_red}{HTML}{D73027}
\definecolor{spec_yellow}{HTML}{FEE08B}
\definecolor{spec_green}{HTML}{1A9641}

\newcommand{\heatmaptext}[4]{%
    \pgfmathsetmacro{\minval}{#2}%
    \pgfmathsetmacro{\maxval}{#3}%
    \pgfmathsetmacro{\value}{#1}%
    \pgfmathsetmacro{\percent}{(\value - \minval) / (\maxval - \minval) * 100}%
    \if#4H
        \pgfmathsetmacro{\colorintensity}{\percent}%
    \else
        \pgfmathsetmacro{\colorintensity}{100 - \percent}%
    \fi%
    
    \ifdim\colorintensity pt < 50pt
        \pgfmathsetmacro{\subintensity}{\colorintensity * 2}%
        {\textcolor{spec_yellow!\subintensity!spec_red}{#1}}%
    \else
        \pgfmathsetmacro{\subintensity}{(\colorintensity - 50) * 2}%
        {\textcolor{spec_green!\subintensity!spec_yellow}{#1}}%
    \fi%
}

\newcommand{\printThreeColorSpectrum}{%
    \begin{tikzpicture}[baseline=0cm]
        \node[anchor=east, font=\footnotesize, inner xsep=2pt] at (0, 0.1cm) {Worst};

        \shade[left color=spec_red, right color=spec_yellow] (0,0) rectangle (2cm, 0.2cm);
        \shade[left color=spec_yellow, right color=spec_green] (2cm,0) rectangle (4cm, 0.2cm);

        \node[anchor=west, font=\footnotesize, inner xsep=2pt] at (4cm, 0.1cm) {Best};
    \end{tikzpicture}%
}

\begin{document}

\title{
Fitts' List Revisited: An Empirical Study on Function Allocation in a Two-Agent Physical Human-Robot Collaborative Position/Force Task}

\author{Nicky Mol\,\orcidlink{0000-0002-2141-9519}, \textit{Graduate Student Member, IEEE},
J. Micah Prendergast\,\orcidlink{0000-0002-9888-3133},
David A. Abbink\,\orcidlink{0000-0001-7778-0090}, \textit{Senior Member, IEEE}, and
Luka Peternel\,\orcidlink{0000-0002-8696-3689} \textit{Member, IEEE}
\thanks{Received 7 May 2025; accepted 25 October 2025. Date of publication 14 November 2025. This article was recommended for publication by Associate Editor Arash Arami and Editor Haoyong Yu upon evaluation of the reviewers’ comments. This work was supported by BrightSky Project, funded by the R\&D Mobiliteitsfonds from the Netherlands Enterprise Agency (RVO) and commissioned by the Ministry of Economic Affairs and Climate Policy. \textit{(Corresponding author: Nicky Mol.)}}%
\thanks{This work involved human subjects or animals in its research. Approval of all ethical and experimental procedures and protocols was granted by the TU Delft Human Research Ethics Committee (HREC) under Application No. 2721, and performed in line with the Declaration of Helsinki.}%
\thanks{Nicky Mol, J. Micah Prendergast, and Luka Peternel are with the Department of Cognitive Robotics, Faculty of Mechanical Engineering, Delft University of Technology, 2628 CD Delft, The Netherlands (e-mail: nicky.mol@tudelft.nl; j.m.prendergast@tudelft.nl; l.peternel@tudelft.nl).}%
\thanks{David A. Abbink is with the Department of Cognitive Robotics, Faculty of Mechanical Engineering, Delft University of Technology, 2628 CD Delft, The Netherlands, and also with the Department of Sustainable Design Engineering, Faculty of Industrial Design Engineering, Delft University of Technology, 2628 CD Delft, The Netherlands (e-mail: d.a.abbink@tudelft.nl).}%
\thanks{This article has supplementary downloadable material available at https://doi.org/10.1109/LRA.2025.3632607, provided by the authors.}%
\thanks{Digital Object Identifier 10.1109/LRA.2025.3632607}%
}

\markboth{IEEE ROBOTICS AND AUTOMATION LETTERS, VOL. 11, NO. 1, JANUARY 2026}
{Mol \MakeLowercase{\textit{et al.}}: Fitts' List Revisited: An Empirical Study on Function Allocation}


\maketitle

\begin{abstract}
In this letter, we investigate whether classical function allocation---the principle of assigning tasks to either a human or a machine---holds for physical Human-Robot Collaboration, which is important for providing insights for Industry 5.0 to guide how to best augment rather than replace workers. This study empirically tests the applicability of Fitts' List within physical Human-Robot Collaboration, by conducting a user study (N=26, within-subject design) to evaluate four distinct allocations of position/force control between human and robot in an abstract blending task. We hypothesize that the function in which humans control the position achieves better performance and receives higher user ratings. When allocating position control to the human and force control to the robot, compared to the opposite case, we observed a significant improvement in preventing overblending. This was also perceived better in terms of physical demand and overall system acceptance, while participants experienced greater autonomy, more engagement and less frustration. An interesting insight was that the supervisory role (when the robot controls both position and force) was rated second best in terms of subjective acceptance. Another surprising insight was that if position control was delegated to the robot, the participants perceived much lower autonomy than when the force control was delegated to the robot. These findings empirically support applying Fitts' principles to static function allocation for physical collaboration, while also revealing important nuanced user experience trade-offs, particularly regarding perceived autonomy when delegating position control.
\end{abstract}


\section{Introduction}
\IEEEPARstart{T}{he} integration of robots in industrial settings has been growing rapidly in recent years, driven by the need to address complex societal challenges such as labor shortages, aging populations, and ever-increasing production demands. While traditional industrial robots have proven their worth in highly controlled and predictable, high-volume settings like automotive assembly lines, they lack the flexibility required for dynamic environments prevalent in high-mix, low-volume manufacturing or maintenance and repair. As a response, collaborative robots (cobots) have emerged, designed specifically for safe operation near human workers. This unlocks the potential for direct physical Human-Robot Collaboration (pHRC) that promises to merge the complementary skills of both agents, creating more effective, flexible, and meaningful collaborations~\cite{ajoudani_progress_2018}. Such vision aligns with the human-centric principles of Industry 5.0, which emphasizes technology in service of worker well-being and resilient production systems.
However, despite the potential of collaborative robots, achieving truly effective, fluent, and satisfying collaboration remains a significant challenge~\cite{michaelis_collaborative_2020}.

\begin{figure}[!t]
 \centering
 \subfloat{\includegraphics[width=0.49\linewidth]{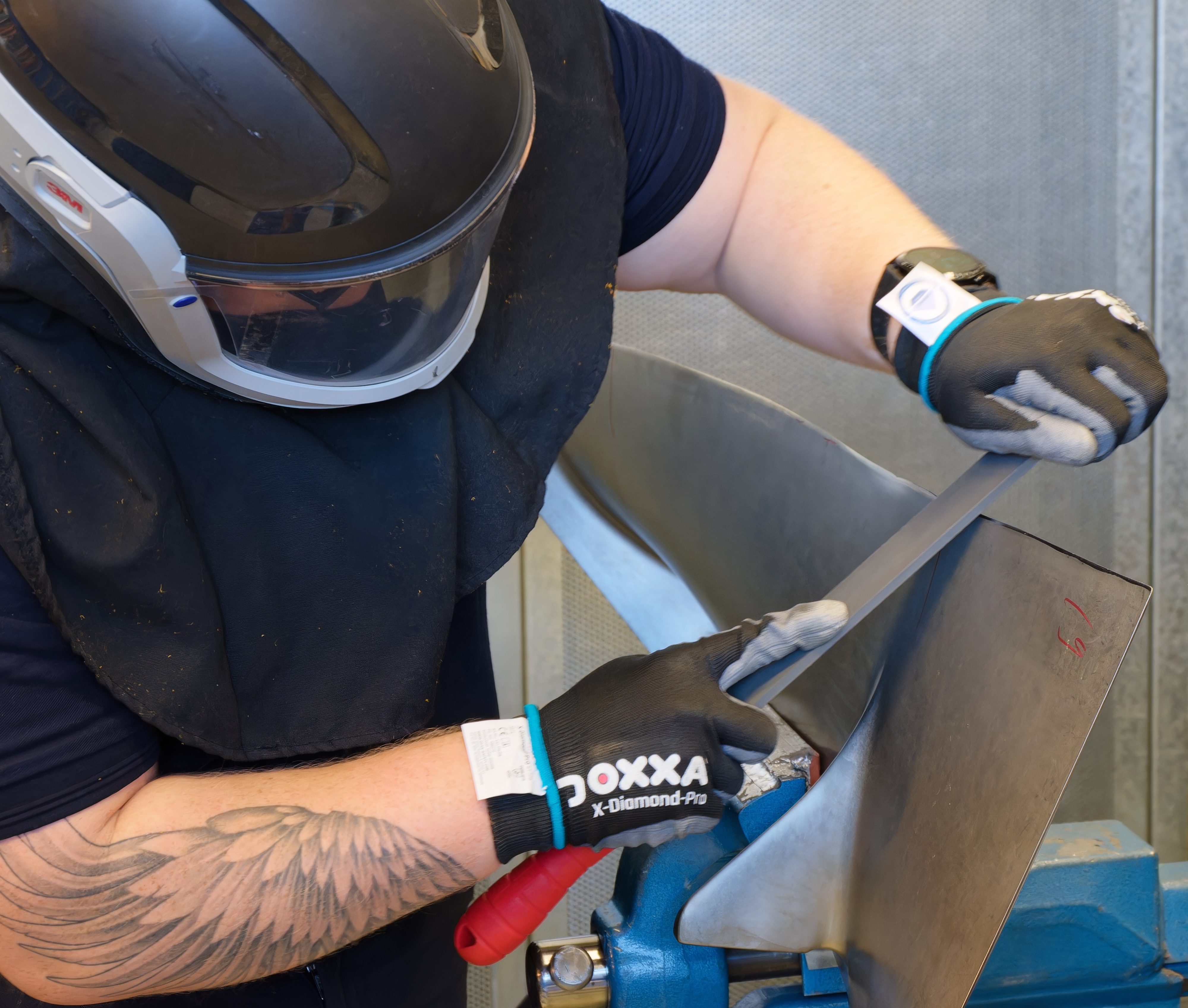}%
 \label{fig:manual_blending}}
 \hfill
 \subfloat{\includegraphics[width=0.49\linewidth]{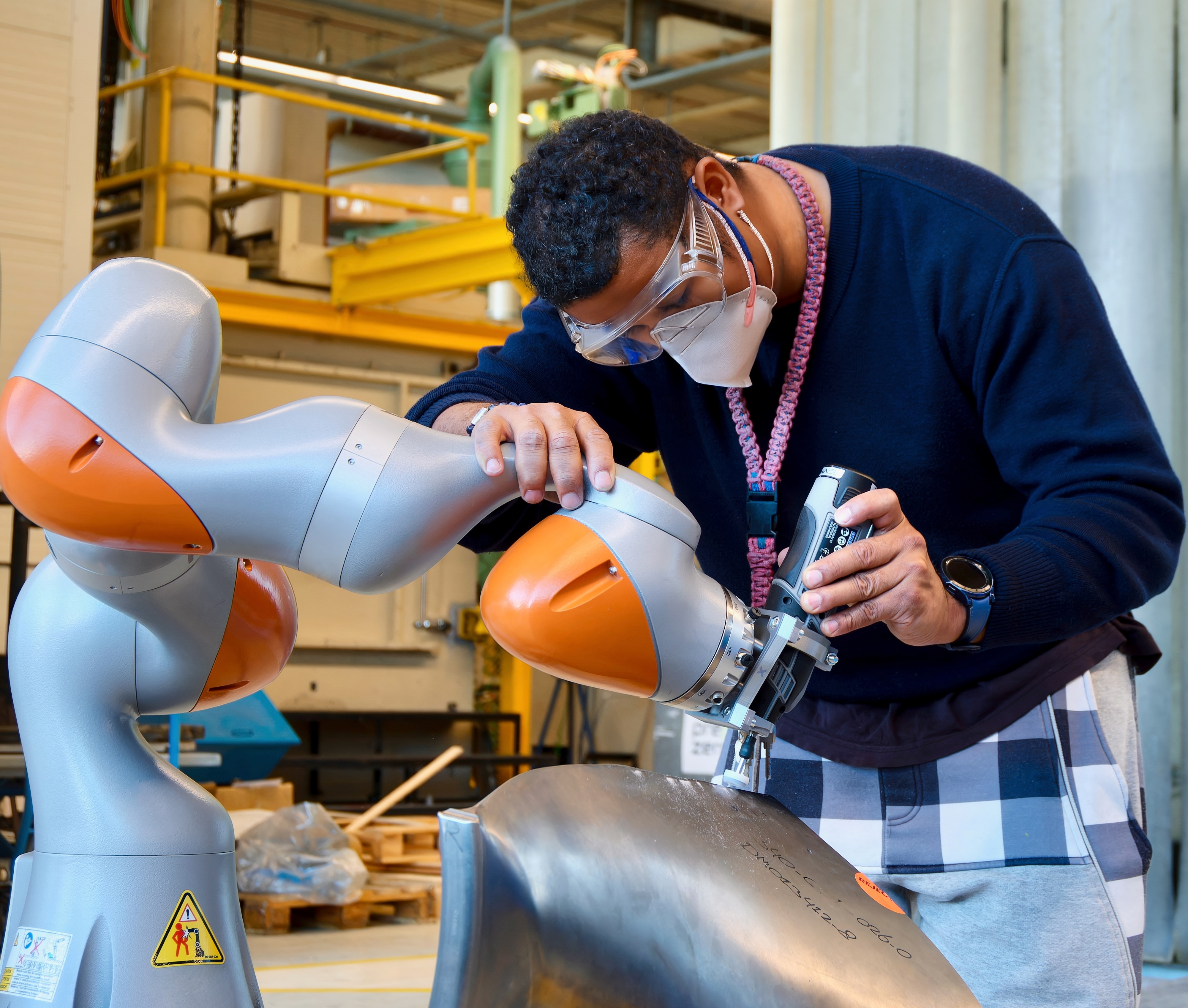}%
 \label{fig:robot_assisted_blending}}
 \caption{Illustrative comparison of a manual blending task (left) versus a collaborative robot assisting a human worker (right) by potentially handling force or position control, based on function allocation.}
 \label{fig:blending_comparison}
 \vspace{-4mm}
\end{figure}

One element of this challenge lies in \textit{function allocation}: deciding which agent--human or robot--should perform which aspects of a shared task~\cite{mortl_role_2012,jarrasse_slaves_2014,el_makrini_hierarchical_2022,selvaggio_autonomy_2021,merlo_ergonomic_2023,vianello_effects_2024}. We can model physical interactions with impedance, which deals with two key variables: motion and forces~\cite{hogan_impedance_1984}. Consider a task like manual blending or polishing of a complex surface, such as an aircraft fan blade (see Fig.~\ref{fig:blending_comparison} left). This requires both precise positioning of the tool and careful application of force. Classic frameworks, notably Fitts' ``MABA-MABA'' (Men Are Better At / Machines Are Better At) list stemming from early human factors research~\cite{fitts_human_1951}, provide initial guidance. Despite its age and extensive criticism, this framework has shown remarkable persistence throughout the history of function allocation~\cite{de_winter_why_2014}. The framework suggests humans generally excel at tasks requiring perception, judgment, and adaptability (e.g., identifying and deciding where to polish), while machines excel at speed, consistent force application, and repetitive precision movements (e.g., producing a constant force). Modern cobots offer the technical capability to split control of force and position between the human and robot (see Fig.~\ref{fig:blending_comparison} right).

Some existing work in pHRC has exploited this ability of cobots to take human inputs during physical interaction, and allocate positions and force sub-tasks. The approach in~\cite{cherubini_collaborative_2016} made the distribution of functions during an assembly task, where the robot performed most of the large movements, while the human stepped in for precise movements. Humans can also step in with physical interaction to give the robot occasional hints on how to re-plan longer movement trajectories in real-time during assembly, where there are multiple possible goals~\cite{haninger_model_2023}. The method in~\cite{roveda_human-robot_2018} was in line with the classical framework, where the assembly of bulky objects was handled by allocating load carrying to the robot, while the precise positioning of the object during installation was allocated to the human. In~\cite{peternel_robot_2018}, a collaborative polishing task was solved by allocating force control to the robot and position control to the human, consistent with Fitts’ principle. In industrial tasks like sanding, letting a robot perform repetitive high-force motions can reduce human physical strain, but fully autonomous solutions often struggle with quality and adaptability~\cite{konstant_humanrobot_2025}. Nevertheless, the existing solutions built on the general Fitts' principle without having a clear insight into whether it is valid in the context of pHRC, prompting a re-evaluation of these classic principles in the context of direct physical collaboration.

\begin{figure*}[t!]
    \centering    
    \includegraphics[width=0.95\textwidth]{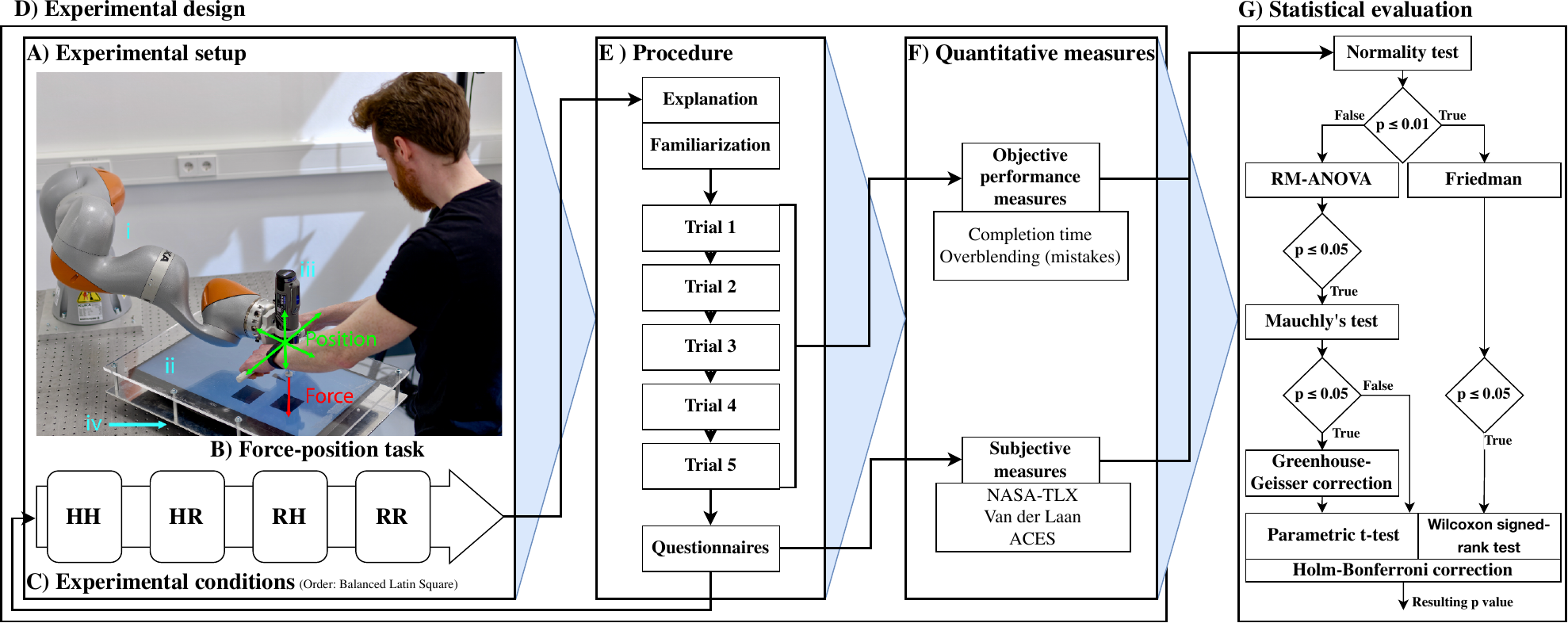}
    \caption{Illustration of pipeline for the method. The experimental setup on the left involves a blending task with force (red) and position (green) sub-tasks. The combinations of assigning these sub-tasks between human (H) and robot (R) result in four experimental conditions. In the labeling of conditions, the first letter is for the position sub-task, while the second letter is for the force sub-task (e.g., the human doing positioning and the robot producing force would be HR). The experiment procedure is followed by the calculation of measures, which are finally used in the statistical evaluation.}
    \label{fig:method}
    \vspace{-4mm}
\end{figure*}

To address this gap and provide the missing insight, this study investigates a fundamental question: \textit{who should control what in a two-agent physical human-robot collaborative position/force task?} To answer this question, we conducted human factors experiments involving a blending task requiring both force application and tool positioning. Participants experienced four distinct conditions representing the extremes of static task allocation on the shared control spectrum: HH (fully manual control), HR (human controls position, robot controls force), RH (robot controls position, human controls force), and RR (robot controls both, human supervises). By collecting and statistically analyzing both objective and subjective data across these conditions, we aimed to gain new insights into function allocation in the context of pHRC. We hypothesize that:
\begin{itemize}
  \item[H1] Van der Laan usefulness and satisfaction score higher in \textbf{HR} than in RH.
  \item[H2] Task performance, in terms of (a) completion time and (b) overblending, score higher in \textbf{HR} than in RH.
  \item[H3] Perceived workload is lower in \textbf{HR} than in RH.
  \item[H4] Perceived autonomy, engagement, competence, and usefulness score higher in \textbf{HR} than in RH.
\end{itemize}

The contributions of this paper are: 1) validating whether Fitts' ``MABA-MABA'' principle holds for pHRC involving hybrid position/force tasks, and 2) providing new user experience insights about autonomy, competence, engagement, and usefulness in different position/force distributions.


\section{Methods}
The method is summarized by a schematic in Fig.~\ref{fig:method}. The experimental setup included a sensorized work surface with an interactive screen (A) for participants to be able to perform a blending task (B). There were four experimental conditions based on human-robot allocation of position and force sub-tasks (C). The procedure involved five trials per condition, and each condition included a brief explanation and familiarization before and questionnaires after the trials (E). Based on the gathered data, both objective and subjective measures were extracted (F) and fed to the statistical evaluation process (G). Each component (A-G) is described separately in detail in the remainder of this section.
 
\subsection{Experimental Setup}
The experimental setup used in this study is illustrated in the leftmost block in Fig.~\ref{fig:method}. It consists of a i) 7-DOF KUKA LBR iiwa 14 R820 (KUKA AG, Augsburg, Germany) robotic manipulator, equipped with a ii) custom mount with handlebars, housing the rotary tool at its end-effector for interacting with the environment. To present the task, a iii) 23.8-inch horizontally aligned Full HD Dell monitor (Dell Technologies Inc., Round Rock, Texas, US) is used. It is encased in a box with a transparent plexiglass surface, which is mechanically connected to a iv) 6-DOF SCHUNK FTN-Delta SI-330-30 Force/Torque sensor (SCHUNK GmbH \& Co. KG, Lauffen/Neckar, Germany), enabling measurement of interaction forces applied to the surface of the box.

The robot is controlled at the joint-torque level using a Cartesian Impedance Controller operating at 200~Hz. The interaction force/torque in Cartesian space was defined as:
\begin{equation}
   \bm{f} = \begin{bmatrix}
   \bm{K}_t & 0\\
    0 & \bm{K}_r
\end{bmatrix}(\bm{x}_a-\bm{x}_d) + \bm{D}(\dot{\bm{x}}_a-\dot{\bm{x}}_d)+\bm{S}_f \bm{f}_{d},
\label{eq:interaction_force}
\end{equation}
where $\bm{K}_t \in \mathbb{R}^{3 \times 3}$ and $\bm{K}_r \in \mathbb{R}^{3 \times 3}$ are the translational and rotational parts, respectively, of the stiffness matrix $\bm{K} \in \mathbb{R}^{6 \times 6}$. $\bm{D} \in \mathbb{R}^{6 \times 6}$ represents the damping matrix, $\bm{x}_a \in \mathbb{R}^{6}$ and $\bm{x}_d \in \mathbb{R}^{6}$ being the actual and desired robot end-effector pose in Cartesian space, $\bm{S}_f \in \mathbb{R}^{6 \times 6}$ a diagonal matrix used for selecting the axis in which the desired force is applied and $\bm{f}_d \in \mathbb{R}^{6}$ the desired force, both in Cartesian space. We used a critically damped design $\bm{D} = 2\zeta \sqrt{\bm{K}}$, with damping ratio $\zeta = 0.7$. The robot and tool were gravity-compensated, while a nullspace controller enforced an ``elbow up'' configuration.

The trajectory generator for reference pose is defined as:
\begin{equation}
   \bm{x}_d(t) = \text{Interp}\left( \bm{x}_{\text{start}}, \bm{x}_{\text{goal}}, \frac{t}{t_\text{total}} \right), \quad \text{for } t \in [0, t_\text{total}],
\label{eq:trajectory}
\end{equation}
where $\bm{x}_d(t)$ is the desired end-effector pose at time $t$, $\text{Interp}()$ is a time-based interpolation function (linear for position and spherical for orientation) taking the start pose ($\bm{x}_{\text{start}}$), goal pose ($\bm{x}_{\text{goal}}$) and the time $t$ divided by the total execution time ($t_\text{total}$), which calculated as follows:
$t_\text{total} = \max \left(\frac{\|\bm{x}_{\text{goal}} - \bm{x}_{\text{start}}\|}{\bm{v}_{\text{d}}} \right)$, with $\bm{v}_{\text{d}}$ being the desired velocity.

\subsection{Position/Force Task}
Participants were tasked with removing damage from a surface, visualized as black rectangles of varying sizes, as shown in the leftmost block in Fig.~\ref{fig:method}. The task can roughly be divided into two sub-tasks: 1) positioning of the tooltip, being held by the robot end-effector, over the damage to be removed and 2) applying a downward force perpendicular to the surface of the monitor to remove the damage. The removal rate is directly related to the applied force (within thresholds) and can be expressed as a function of the applied force:
\begin{equation}
    \frac{\partial h}{\partial t}(f_{\text{z}}) = 
    \begin{cases} 
        0, & \text{if } f_{\text{z}} < f_{\text{min}} \\
        \displaystyle \frac{h_{\text{max}}}{t_{\text{min}}} \cdot \frac{f_{\text{z}} - f_{\text{min}}}{f_{\text{max}} - f_{\text{min}}}, & \text{if } f_{\text{min}} \leq f_{\text{z}} \leq f_{\text{max}} \\
        \displaystyle \frac{h_{\text{max}}}{t_{\text{min}}}, & \text{if } f_{\text{z}} > f_{\text{max}}
    \end{cases},
    \label{eq:removal_rate}
\end{equation}
where $\frac{\partial h}{\partial t}$ is the removal rate, defined as the amount of health points removed per second. $f_{\text{z}}$ is the force applied by the participant along the z-axis perpendicular to the surface as measured by the force/torque sensor, $f_{\text{min}}$ and $f_{\text{max}}$ are the minimum and maximum force thresholds. Forces below $f_{\text{min}}$ do not contribute to removal, while forces above $f_{\text{max}}$ saturate the removal rate. $h_{\text{max}}$ is the maximum amount of health points per pixel, and $t_{\text{min}}$ is defined to be the minimum time required to completely remove a pixel with $h_{\text{max}}$ when applying the maximum effective force $f_{\text{max}}$.

Each trial involved the removal of three rectangular damage blocks consisting of pixels with an initial amount of health points. As the task progressed, damage opacity decreased proportionally to the remaining health value, providing continuous visual feedback. A rectangular block was considered successfully removed when at least $95\%$ of its initial health had been removed, at which point the block turns green and fades from view, indicating its completion. When participants applied force to an area where no damage (or no remaining damage) was present, they introduced new damage (called overblending) to which the same force-dependent removal model (Eq.~\eqref{eq:removal_rate}) was applied and was visually indicated by an increase in opacity and a color shift from black to red of the affected pixels.

\subsection{Experimental Conditions}
The study included four experimental conditions that represent all combinations of extremes on the shared control spectrum for both the position and force sub-tasks. The conditions are summarized in Table~\ref{tab:conditions}, where also the differences in terms of control with respect to Eq.~\eqref{eq:interaction_force} are specified.
\begin{table}[h]
    \centering
    \caption{Conditions and their differences in terms of control}
    \begin{tabular}{lcccc}
        \toprule
        \textbf{Condition} & \textbf{Position} & \textbf{Force} & $\bm{K}_t$ & $f_\text{d,z}$ \\
        \midrule
        1: HH & Human & Human & $\bm{K}_t = \bm{0}$ & $f_\text{d,z} = 0$  \\
        2: HR & Human & Robot & $\bm{K}_t = \bm{0}$ & $f_\text{d,z} = f_{\text{max}}$ \\
        3: RH & Robot & Human & $\bm{K}_t \neq \bm{0}$ & $f_\text{d,z} = 0$ \\
        4: RR & Robot & Robot & $\bm{K}_t \neq \bm{0}$ & $f_\text{d,z}  = f_{\text{max}}$
        \\
        \bottomrule
    \end{tabular}
    \label{tab:conditions}
\end{table}

\textbf{Condition 1 - HH}: Participants control both the position of the tool as well as the application of force.

\textbf{Condition 2 - HR}: Participants control the position of the tool while the robot applies force.

\textbf{Condition 3 - RH}: Robot controls the position of the tool while participants apply force.

\textbf{Condition 4 - RR}: Robot controls both the position of the tool as well as the application of force, while participants supervise and intervene in case the robot makes mistakes.

\subsection{Experimental Design}
In a within-subject design, 26 participants (aged 22–42, 4 female, 22 male, physically healthy) completed the task under all four conditions. The study was approved by the TU Delft Human Research Ethics Committee. All participants gave written informed consent prior to their participation. To mitigate order effects, a Balanced Latin Square design was applied on the order in which the conditions were presented to the participants. Each trial involved removing three damage spots under consistent layout constraints to control task difficulty across conditions.

The four conditions were designed to isolate the effects of delegating position and/or force control: a manual baseline (HH), an automated supervisory condition (RR), and two ``mixed-task'' conditions (HR and RH). RR was tuned to operate near the performance limits of what is physically possible for a human to achieve. All conditions used the same Cartesian impedance controller in Eq.~\eqref{eq:interaction_force}, with condition-specific translational stiffness settings. For HH and HR, translational elements of the stiffness matrix $\bm{K}$ were set to $0$ N/m in all directions, to allow for unhindered human positioning. For RH, it was tuned to $5000$ N/m in $x$ and $y$ and for RR in all translational directions. Rotational stiffness was fixed at $100$ Nm/rad in all conditions. These values were determined through pilot testing to be high enough to rigidly guide the user's motion along the robot's path without causing instability. Force thresholds were set at $f_{\text{min}} = 5$ N and $f_{\text{max}} = 40$ N. Pixel damage removal time at $f_{\text{max}}$ was $t_{\text{min}} = 2$ s, with $h_{\text{max}} = 100$ health points per pixel. Trajectory velocities (RH/RR) were tuned to $0.0083$ m/s (blending), $0.0400$ m/s ((un)latching), and $0.0800$ m/s (travel).

In HR, the robot applied a force capped at $f_{\text{max}}$ whenever the tooltip was within 1~mm of the surface and stopped when it moved above or below 3~mm, creating an intuitive latching effect consistent across participants. In RR, the robot autonomously controlled both position and force by following a predefined adaptive trajectory that systematically covered the damage spot, divided into sections of equal width. If insufficient removal was detected in a section, the robot would revisit it. To make the task engaging for the human in the supervisory role, at fixed points, automation errors were introduced by deviating the trajectory outside the damaged area, leading to overblending unless the participant intervened promptly by pressing a button. This condition modeled supervisory control, acknowledging that full autonomy is infeasible in most complex practical tasks~\cite{bradshaw_seven_2013}.

\subsection{Procedure}
Participants first received a general explanation of the experiment and task, where they were instructed to remove three damage spots as quickly as possible while minimizing overblending. Before each condition, they were briefed on its specifics and completed a 2-minute familiarization period. They then performed five trials, where damage spots were always the same size and remained at the same pair‑wise distance from each other, but were rotated, translated, or flipped. After the trials, the participants were asked to fill out three questionnaires.

\subsection{Quantitative Measures}
The objective performance metrics that were considered are the time of completion, which is defined as the amount of time it takes the participant to remove the damage during a trial, and overblended health defined as the amount of cumulative health points resulting from applying force at locations where there is no damage (any longer). Participants' perceived acceptance was assessed using the Van der Laan usefulness and satisfying scales~\cite{van_der_laan_simple_1997}, while perceived workload was assessed using the NASA-TLX questionnaire~\cite{hart_nasa-task_2006}. 

To probe the psychological impact of pHRC beyond standard metrics~\cite{fletcher_we_2023}, we developed the exploratory Agentic Collaboration and Engagement Scale (ACES) because existing scales from workplace psychology are ill-suited for a single-session, abstracted experiment and lack a human-robot role perspective. 
The scale’s theoretical foundation draws from self-determination theory (SDT), which identifies \textbf{autonomy} and \textbf{competence} as basic psychological needs that foster intrinsic motivation and \textbf{engagement}~\cite{gagne_understanding_2022}. These factors, together with a sense of \textbf{usefulness}, are drivers of meaningful work, an experience that is known to enhance job satisfaction and well-being~\cite{smids_robots_2020}.
To assess reliability, Cronbach's alpha ($\alpha$) was calculated, which showed an internal consistency of $\alpha=0.73$, 95\% CI [0.63, 0.80] for the ``Human'' scale (measuring perceived meaningful involvement) and $\alpha=0.79$, 95\% CI [0.71, 0.85] for the ``Robot'' scale (measuring perceived quality of the robot teammate).
ACES uses a dual-perspective format, rating each construct for the participant's experience (``I'') and their perception of the robot (``the robot'') on a Likert scale using four statements:
\begin{itemize}
\item I felt that XX had a high level of autonomy while completing the task.
\item I felt that XX had the necessary skills and knowledge to complete the part of the task.
\item I felt XX was useful in performing its part of the task.
\item I felt XX was engaged in the role.
\end{itemize}

\subsection{Statistical Evaluation}
As shown in Fig.~\ref{fig:method} G), data normality was assessed using the Shapiro–Wilk test. If normality was met ($p > 0.01$, including minor violations), a Repeated Measures ANOVA was used. Violations of sphericity (Mauchly’s test) were corrected with Greenhouse–Geisser. Significant main effects were followed by paired $t$-tests with Holm–Bonferroni correction. Effect size was reported as $\eta_{p}^{2}$.
If normality was majorly violated ($p \leq 0.01$), a Friedman test was applied. Post-hoc Wilcoxon signed-rank tests with Holm–Bonferroni correction were used for pairwise comparisons, with effect size reported using Kendall’s W.


\section{Results}
Here, the experimental results are presented, highlighting the differences across the four conditions (HH, HR, RH, RR). Findings are structured by measure type, referencing Figures~\ref{fig:vdl}-\ref{fig:mol-cis} and the detailed statistical evaluation in Table~\ref{tab:statistical_analysis}.

\subsection{Van der Laan System Acceptance}
\begin{figure}
    \centering
    \includegraphics[width=0.8\linewidth]{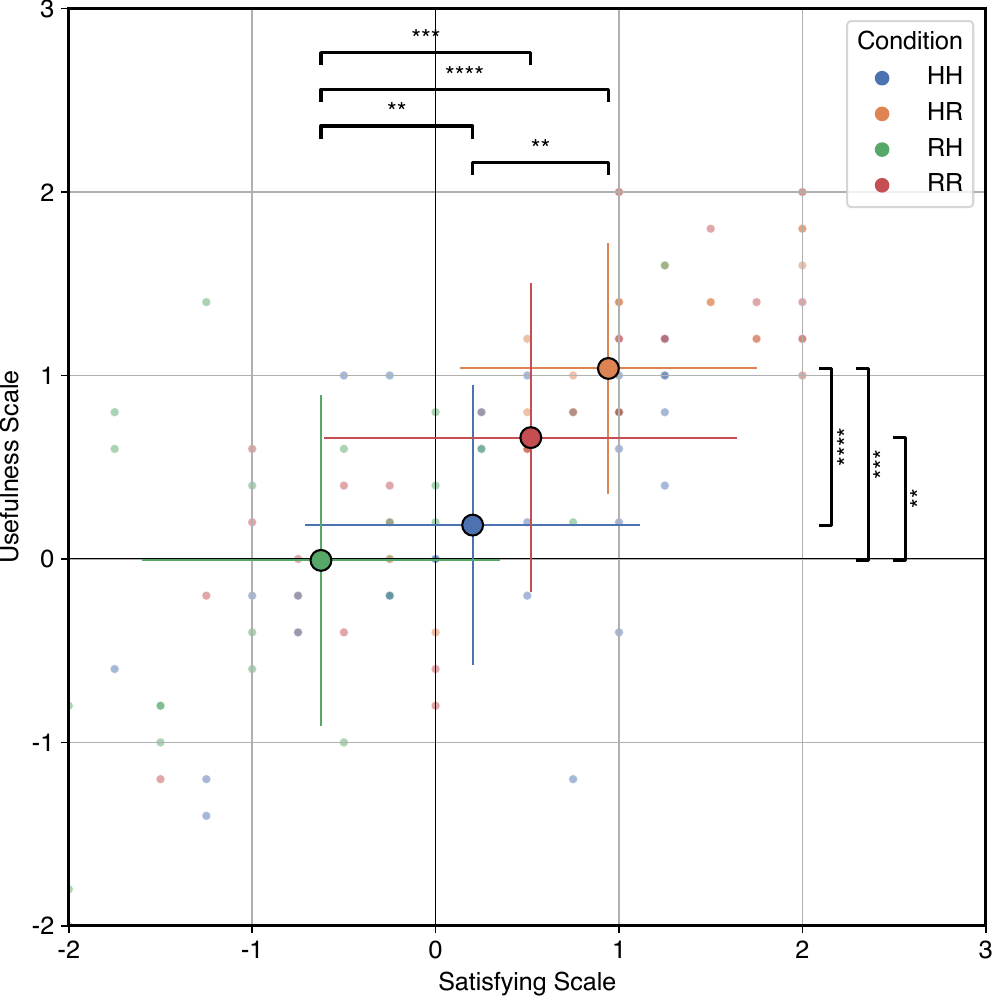}
    \caption{Plot showing subjective scores on the Van der Laan acceptance scales: usefulness (y‑axis) versus satisfying (x‑axis). Large circles with error bars show the mean $\pm$ SD for each condition, while small semi‑transparent dots are the corresponding single‑participant scores. Points that lie higher and further right indicate greater perceived acceptance. Statistically significant pair‑wise differences are annotated on each axis, where: **:~$0.001 < p \le 0.01$; ***:~$0.0001 < p \le 0.001$; ****:~$p \le 0.0001$.}
    \label{fig:vdl}
    \vspace{-4mm}
\end{figure}

Acceptance was evaluated using the Van der Laan questionnaire. Figure~\ref{fig:vdl} shows a scatter plot of the results, displaying individual participant ratings along with the mean scores and standard deviation error bars for each condition, annotated with statistically significant pairwise comparisons.

Significant differences were found across conditions for perceived usefulness ($\text{F}(2.7, 66.7) = 12.39\text{, }p < 0.0001$) and satisfaction ($\text{F}(2.3, 56.9) = 13.55\text{, }p<0.0001$). Pairwise comparisons revealed a strong preference for HR over RH, with HR rated significantly higher on both usefulness (t-test, $p=0.0002$) and satisfaction (t-test, $p<0.0001$). HR was also rated significantly more useful (t-test, $p<0.0001$) and satisfying (t-test, $p=0.0028$) than the baseline HH condition. Conversely, RH showed no usefulness benefit over HH (t-test, $p=0.2810$) and was significantly less satisfying (t-test, $p=0.0029$). Notably, the supervisory RR condition was rated significantly higher than RH on usefulness (t-test, $p=0.0086$) and satisfaction (t-test, $p=0.0009$), positioning it second in user acceptance, statistically comparable to HH but generally lower than HR.

\subsection{NASA-TLX Workload}
\begin{figure}
    \centering
    \includegraphics[width=0.9\linewidth]{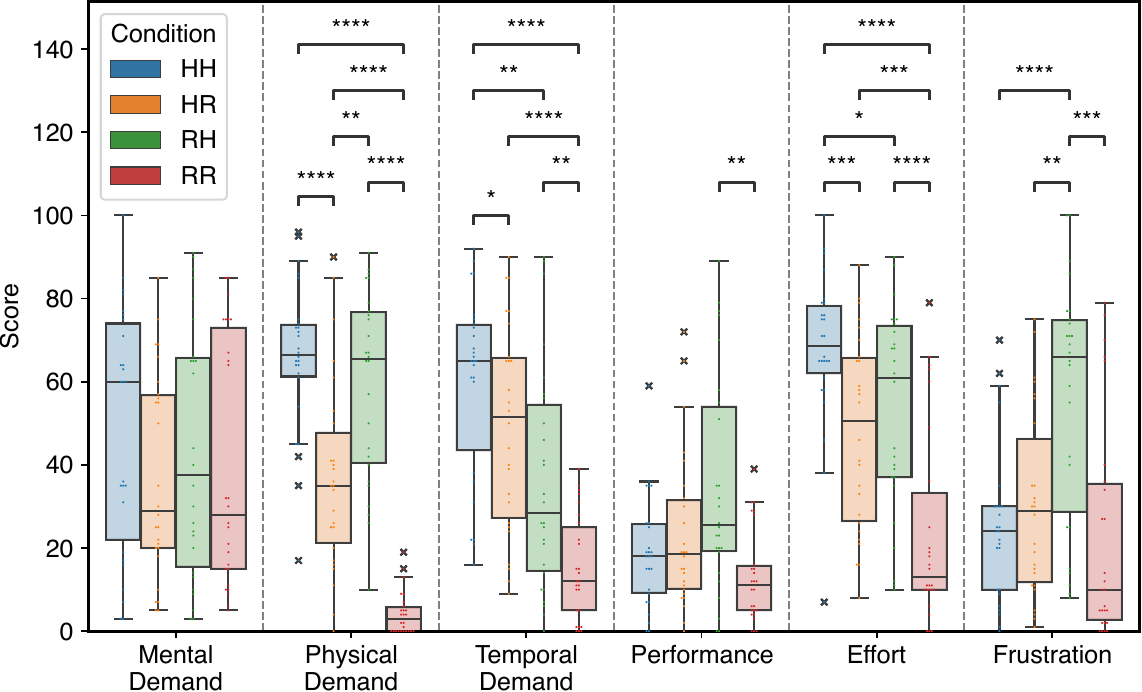}
    \caption{Box plots showing NASA‑TLX scores. Each swimlane shows the results of one workload sub‑scale for each of the conditions. Boxes show the median and the interquartile range, where whiskers extend to the 5th and 95th percentiles, while small dots are the individual participant scores. Higher scores correspond to greater perceived workload. Statistically significant pair‑wise differences are annotated, where: *:~$0.01 < p \le 0.05$; **:~$0.001 < p \le 0.01$; ***:~$0.0001 < p \le 0.001$; ****:~$p \le 0.0001$.}
    \label{fig:nasa-tlx}
    \vspace{-4mm}
\end{figure}

Workload was evaluated using the NASA-TLX questionnaire. Box-plots showing median and interquartile range of the scores for each condition per item are visualized in Fig.~\ref{fig:nasa-tlx}, annotated with statistically significant pairwise comparisons.

Significant effects were found for all workload dimensions except mental demand ($\text{F}(3, 75)=1.19\text{, } p=0.3207$) (Fig.~\ref{fig:nasa-tlx}). Physical demand showed the largest differences ($\text{F}(2.3, 57.6)=63.35\text{, }p<0.0001$), with RR being significantly lowest (t-test, $p<0.0001$) compared to all other conditions. HR significantly reduced physical demand compared to both HH (t-test, $p<0.0001$) and RH (t-test, $p=0.0039$). Effort followed similar patterns, being lowest in RR. Frustration was uniquely high in the RH condition, significantly higher than HH (t-test, $p<0.0001$), HR (t-test, $p=0.0040$), and RR (t-test, $p=0.0002$). Although mental demand showed no significant main effect, variability within conditions was high (see SDs in Table \ref{tab:statistical_analysis}).

\subsection{Performance Metrics}
Performance was evaluated using two objective metrics: task completion time and amount of overblending as a measure of mistakes made. Box-plots showing median and interquartile range of the scores for each condition are visualized in Fig.~\ref{fig:performance}, annotated with statistically significant pairwise comparisons.

Both completion time ($\chi^2(3) = 49.71\text{, }p < 0.0001$) and overblended health ($\chi^2(3) = 16.34\text{, }p = 0.0010$) showed significant differences. RR yielded significantly faster completion times than all other conditions (Wilcoxon signed-rank, $p<0.0001$), while overblended health was significantly higher in RH compared to HR (Wilcoxon signed-rank, $p=0.0397$) and RR (Wilcoxon signed-rank, $p<0.0001$).

\subsection{Agentic Collaboration and Engagement Scale (ACES)}
The subjective quality of the collaboration in terms of autonomy, competence, usefulness, and engagement was evaluated using ACES based on a Likert scale and was rated on how the participants perceived themselves and how they perceived the robot. Median and interquartile ranges per condition are shown in Fig.~\ref{fig:mol-cis}.

Significant differences were observed for participants' ratings of their own autonomy ($\chi^2$(3) = 64.78, $p<0.0001$), competence ($\chi^2$(3) = 17.83, $p=0.0005$, usefulness ($\text{F}(2.1, 52.8)=34.63\text{, } p<0.0001$), and engagement ($\text{F}(2.1, 53.0)=15.65\text{, } p<0.0001$).
Ratings for autonomy, usefulness, and engagement were highest when participants controlled position (HH and HR). Specifically, HR yielded significantly higher ratings compared HR on autonomy (Wilcoxon signed-rank, $p<0.0001$), usefulness (t-test, $p<0.0001$), and engagement (t-test, $p=0.0122$). Delegating position reduced perceived autonomy significantly more (HH vs RH: Wilcoxon signed-rank, $p<0.0001$) than delegating force (HH vs HR: Wilcoxon signed-rank, $p=0.0034$). Self-rated competence was lowest in RH, significantly lower than in RR (Wilcoxon signed-rank, $p=0.0012$).

Significant differences were also observed for ratings of the robot's autonomy ($\chi^2$(3) = 60.75, $p<0.0001$), usefulness ($\text{F}(2.1, 52.0)=15.98\text{, } p<0.0001$), and engagement ($\text{F}(2.0, 50.6)=25.10\text{, } p<0.0001$), except for competence ($\text{F}(2.0, 51.2)=2.27\text{, } p=0.1127$). Robot autonomy was perceived highest when it controlled position (RH and RR). The robot in the RR condition was rated significantly most useful and engaging compared to all other conditions.

\begin{table*}[ht!]
\caption{Reported descriptive and inferential statistics of the qualitative measures}
\footnotesize
\setlength{\tabcolsep}{1.6pt}
\resizebox{0.99\textwidth}{!}{
\begin{tabular}{@{}llcccccccccc@{}}
\hline
\textbf{Measure} &  & \textbf{HH} & \textbf{HR} & \textbf{RH} & \textbf{RR} & \textbf{Main Effect} & \textbf{Effect Size} & \multicolumn{3}{c}{\textbf{Piece-wise Comparison}\textsuperscript{b}} \\
\hline

\textbf{Van der Laan}\textsuperscript{a} \\
\hspace{3mm} Usefulness & M & \heatmaptext{0.18}{-0.01}{1.04}{H} & \heatmaptext{1.04}{-0.01}{1.04}{H} & \heatmaptext{-0.01}{-0.01}{1.04}{H} & \heatmaptext{0.66}{-0.01}{1.04}{H} & F(2.7,66.7)=12.39 & $\eta^2_p$=0.331 & 1: \textbf{p$<$0.0001} & 2: p=0.2810 & 3: p=0.1038 \\
\hspace{3mm} (-2, 2)$\uparrow$ & SD & 0.76 & 0.69 & 0.90 & 0.84 & \textbf{p$<$0.0001}$^{\dagger}$ &  & 4: \textbf{p=0.0002} & 5: p=0.1038 & 6: \textbf{p=0.0086} \\
\hspace{3mm} Satisfying score & M & \heatmaptext{0.20}{-0.62}{0.94}{H} & \heatmaptext{0.94}{-0.62}{0.94}{H} & \heatmaptext{-0.62}{-0.62}{0.94}{H} & \heatmaptext{0.52}{-0.62}{0.94}{H} & F(2.3,56.9)=13.55 & $\eta^2_p$=0.351 & 1: \textbf{p=0.0028} & 2: \textbf{p=0.0029} & 3: p=0.3371 \\
\hspace{3mm} (-2, 2) $\uparrow$ & SD & 0.91 & 0.81 & 0.98 & 1.13 & \textbf{p$<$0.0001}$^{\dagger}$ &  & 4: \textbf{p$<$0.0001} & 5: p=0.2138 & 6: \textbf{p=0.0009} \\
\midrule
\textbf{NASA-TLX} \\
\hspace{3mm} Mental demand & Mdn & \heatmaptext{60.00}{28}{60}{L} & \heatmaptext{29.00}{28}{60}{L}  & \heatmaptext{37.50}{28}{60}{L}  & \heatmaptext{28.00}{28}{60}{L} & F(3,75)=1.19 & $\eta^2_p$=0.045 & 1: p=0.3017 & 2: p=1.0000 & 3: p=1.0000 \\
\hspace{3mm} (0, 100) $\downarrow$ & IQR & 22.00-74.00 & 20.00-56.75 & 15.50-65.75 & 15.00-73.00 & p=0.3207 &  & 4: p=1.0000 & 5: p=1.0000 & 6: p=1.0000 \\
\hspace{3mm} Physical demand & Mdn & \heatmaptext{66.50}{3}{66.5}{L} & \heatmaptext{35.00}{3}{66.5}{L} & \heatmaptext{65.50}{3}{66.5}{L} & \heatmaptext{3.00}{3}{66.5}{L} & F(2.3,57.6)=63.35 & $\eta^2_p$=0.717 & 1: \textbf{p$<$0.0001} & 2: p=0.2603 & 3: \textbf{p$<$0.0001} \\
\hspace{3mm} (0, 100) $\downarrow$ & IQR & 61.25-73.75 & 21.25-47.75 & 40.50-76.75 & 0.00-5.75 & \textbf{p$<$0.0001}$^{\dagger}$ &  & 4: \textbf{p=0.0039} & 5: \textbf{p$<$0.0001} & 6: \textbf{p$<$0.0001} \\
\hspace{3mm} Temporal demand & Mdn & \heatmaptext{65.00}{12}{65}{L} & \heatmaptext{51.50}{12}{65}{L} & \heatmaptext{28.50}{12}{65}{L} & \heatmaptext{12.00}{12}{65}{L} & F(2.0,49.5)=24.41 & $\eta^2_p$=0.494 & 1: \textbf{p=0.0440} & 2: \textbf{p=0.0019} & 3: \textbf{p$<$0.0001} \\
\hspace{3mm} (0, 100) $\downarrow$ & IQR & 43.50-73.75 & 27.25-65.75 & 14.50-54.50 & 5.00-25.00 & \textbf{p$<$0.0001}$^{\dagger}$ &  & 4: p=0.0953 & 5: \textbf{p$<$0.0001} & 6: \textbf{p=0.0019} \\
\hspace{3mm} Performance & Mdn & \heatmaptext{18.00}{11}{25.5}{L} & \heatmaptext{18.50}{11}{25.5}{L} & \heatmaptext{25.50}{11}{25.5}{L} & \heatmaptext{11.00}{11}{25.5}{L} & $\chi^2$(3)=13.04 & W=0.167 & 1: p=0.7878 & 2: p=0.1031 & 3: p=0.1903 \\
\hspace{3mm} (0, 100) $\downarrow$ & IQR & 9.25-25.75 & 10.25-31.50 & 19.25-54.00 & 5.00-15.75 & \textbf{p=0.0046}$^{\circ\circ}$ &  & 4: p=0.1779 & 5: p=0.1031 & 6: \textbf{p=0.0018} \\
\hspace{3mm} Effort & Mdn & \heatmaptext{68.50}{13}{68.5}{L} & \heatmaptext{50.50}{13}{68.5}{L} & \heatmaptext{61.00}{13}{68.5}{L} & \heatmaptext{13.00}{13}{68.5}{L} & F(3,75)=24.17 & $\eta^2_p$=0.492 & 1: \textbf{p=0.0010} & 2: \textbf{p=0.0136} & 3: \textbf{p$<$0.0001} \\
\hspace{3mm} (0, 100) $\downarrow$ & IQR & 62.00-78.25 & 26.50-65.75 & 37.00-73.50 & 10.00-33.25 & \textbf{p$<$0.0001}$^\circ$ &  & 4: p=0.3790 & 5: \textbf{p=0.0004} & 6: \textbf{p$<$0.0001} \\
\hspace{3mm} Frustration & Mdn & \heatmaptext{24.00}{10}{66}{L} & \heatmaptext{29.00}{10}{66}{L} & \heatmaptext{66.00}{10}{66}{L} & \heatmaptext{10.00}{10}{66}{L} & F(3,75)=12.41 & $\eta^2_p$=0.332 & 1: p=0.8882 & 2: \textbf{p$<$0.0001} & 3: p=0.8882 \\
\hspace{3mm} (0, 100) $\downarrow$ & IQR & 10.00-30.00 & 11.75-46.25 & 28.75-74.75 & 2.75-35.50 & \textbf{p$<$0.0001} &  & 4: \textbf{p=0.0040} & 5: p=0.8882 & 6: \textbf{p=0.0002} \\
\midrule
\textbf{Performance metrics} \\
\hspace{3mm} Completion time & Mdn & \heatmaptext{109.99}{90.72}{112.87}{L} & \heatmaptext{112.87}{90.72}{112.87}{L} & \heatmaptext{109.33}{90.72}{112.87}{L} & \heatmaptext{90.72}{90.72}{112.87}{L} & $\chi^2$(3)=49.71 & W=0.637 & 1: p=0.0528 & 2: p=0.4992 & 3: \textbf{p$<$0.0001} \\
\hspace{3mm} (s) $\downarrow$ & IQR & 103.88-115.77 & 105.70-119.33 & 103.12-111.07 & 90.09-92.01 & \textbf{p$<$0.0001}$^{\circ\circ}$ &  & 4: p=0.1261 & 5: \textbf{p$<$0.0001} & 6: \textbf{p$<$0.0001} \\
\hspace{3mm} Overblended health & Mdn & \heatmaptext{0}{0}{12368}{L} & \heatmaptext{21}{0}{12368}{L} & \heatmaptext{12368}{0}{12368}{L} & \heatmaptext{5269}{0}{12368}{L} & $\chi^2$(3)=16.34 & W=0.209 & 1: p=1.0000 & 2: p=0.0705 & 3: p=1.0000 \\
\hspace{3mm} (health points) $\downarrow$ & IQR & 0-11866 & 0-9147 & 3467-36513 & 3993-8156 & \textbf{p=0.0010}$^{\circ\circ}$ &  & 4: \textbf{p=0.0397} & 5: p=1.0000 & 6: \textbf{p$<$0.0001} \\
\midrule
\textbf{ACES - Human} \\
\hspace{3mm} Autonomy & Mdn & \heatmaptext{20.50}{4.5}{20.5}{H} & \heatmaptext{18.50}{4.5}{20.5}{H} & \heatmaptext{6.50}{4.5}{20.5}{H} & \heatmaptext{4.50}{4.5}{20.5}{H} & $\chi^2$(3)=64.78 & W=0.831 & 1: \textbf{p=0.0034} & 2: \textbf{p$<$0.0001} & 3: \textbf{p$<$0.0001} \\
\hspace{3mm} (1, 21) $\uparrow$ & IQR & 19.00-21.00 & 15.00-20.00 & 5.00-7.75 & 3.00-7.75 & \textbf{p$<$0.0001}$^{\circ\circ}$ &  & 4: \textbf{p$<$0.0001} & 5: \textbf{p$<$0.0001} & 6: p=0.6250 \\
\hspace{3mm} Competence & Mdn & \heatmaptext{18.50}{16}{19}{H} & \heatmaptext{18.00}{16}{19}{H} & \heatmaptext{16.00}{16}{19}{H} & \heatmaptext{19.00}{16}{19}{H} & $\chi^2$(3)=17.83 & W=0.229 & 1: p=0.8262 & 2: p=0.0704 & 3: p=0.7311 \\
\hspace{3mm} (1, 21) $\uparrow$ & IQR & 16.25-20.00 & 15.25-20.00 & 11.00-19.00 & 17.00-21.00 & \textbf{p=0.0005}$^{\circ\circ}$ &  & 4: p=0.1076 & 5: p=0.3049 & 6: \textbf{p=0.0012} \\
\hspace{3mm} Usefulness & Mdn & \heatmaptext{19.00}{9}{19}{H} & \heatmaptext{18.00}{9}{19}{H} & \heatmaptext{9.00}{9}{19}{H} & \heatmaptext{9.50}{9}{19}{H} & F(2.1,52.8)=34.63 & $\eta^2_p$=0.581 & 1: p=0.2095 & 2: \textbf{p$<$0.0001} & 3: \textbf{p$<$0.0001} \\
\hspace{3mm} (1, 21) $\uparrow$ & IQR & 18.00-20.00 & 16.00-20.00 & 6.25-15.00 & 7.00-14.00 & \textbf{p$<$0.0001}$^{\dagger}$ &  & 4: \textbf{p$<$0.0001} & 5: \textbf{p$<$0.0001} & 6: p=0.7471 \\
\hspace{3mm} Engagement & Mdn & \heatmaptext{18.50}{12}{18.5}{H} & \heatmaptext{16.00}{12}{18.5}{H} & \heatmaptext{13.00}{12}{18.5}{H} & \heatmaptext{12.00}{12}{18.5}{H} & F(2.1,53.0)=15.65 & $\eta^2_p$=0.385 & 1: \textbf{p=0.0044} & 2: \textbf{p$<$0.0001} & 3: \textbf{p$<$0.0001} \\
\hspace{3mm} (1, 21) $\uparrow$ & IQR & 15.50-20.00 & 13.50-19.00 & 7.00-16.50 & 5.50-14.00 & \textbf{p$<$0.0001}$^{\dagger}$ &  & 4: \textbf{p=0.0122} & 5: \textbf{p=0.0010} & 6: p=0.3431 \\
\midrule
\textbf{ACES - Robot} \\
\hspace{3mm} Autonomy & Mdn & \heatmaptext{3.00}{3}{20}{H} & \heatmaptext{8.00}{3}{20}{H} & \heatmaptext{15.50}{3}{20}{H} & \heatmaptext{20.00}{3}{20}{H} & $\chi^2$(3)=60.75 & W=0.779 & 1: \textbf{p$<$0.0001} & 2: \textbf{p$<$0.0001} & 3: \textbf{p$<$0.0001} \\
\hspace{3mm} (1, 21) $\uparrow$ & IQR & 2.00-4.00 & 5.00-11.75 & 13.00-18.00 & 18.00-21.00 & \textbf{p$<$0.0001}$^{\circ\circ}$ &  & 4: \textbf{p=0.0035} & 5: \textbf{p$<$0.0001} & 6: \textbf{p=0.0027} \\
\hspace{3mm} Competence & Mdn & \heatmaptext{18.50}{16}{18.5}{H} & \heatmaptext{17.50}{16}{18.5}{H} & \heatmaptext{16.00}{16}{18.5}{H} & \heatmaptext{18.00}{16}{18.5}{H} & F(2.0,51.2)=2.27 & $\eta^2_p$=0.083 & 1: p=0.5993 & 2: p=0.7554 & 3: p=0.5074 \\
\hspace{3mm} (1, 21) $\uparrow$ & IQR & 11.00-20.00 & 14.00-19.00 & 13.25-18.00 & 16.00-20.00 & p=0.1127$^{\circ\dagger}$ &  & 4: p=0.4529 & 5: p=0.5074 & 6: p=0.1117 \\
\hspace{3mm} Usefulness & Mdn & \heatmaptext{8.50}{8.5}{19}{H} & \heatmaptext{17.00}{8.5}{19}{H} & \heatmaptext{16.00}{8.5}{19}{H} & \heatmaptext{19.00}{8.5}{19}{H} & F(2.1,52.0)=15.98 & $\eta^2_p$=0.390 & 1: \textbf{p=0.0010} & 2: \textbf{p=0.0155} & 3: \textbf{p$<$0.0001} \\
\hspace{3mm} (1, 21) $\uparrow$ & IQR & 4.00-15.50 & 14.25-19.00 & 12.25-18.75 & 18.00-20.00 & \textbf{p$<$0.0001}$^{\circ\dagger}$ &  & 4: p=0.1807 & 5: \textbf{p=0.0135} & 6: \textbf{p=0.0078} \\
\hspace{3mm} Engagement & Mdn & \heatmaptext{3.00}{3}{20}{H} & \heatmaptext{8.00}{3}{20}{H} & \heatmaptext{15.50}{3}{20}{H} & \heatmaptext{20.00}{3}{20}{H} & F(2.0,50.6)=25.10 & $\eta^2_p$=0.501 & 1: \textbf{p=0.0001} & 2: \textbf{p=0.0003} & 3: \textbf{p$<$0.0001} \\
\hspace{3mm} (1, 21) $\uparrow$ & IQR & 2.00-4.00 & 5.00-11.75 & 13.00-18.00 & 18.00-21.00 & \textbf{p$<$0.0001}$^{\dagger}$ &  & 4: p=0.1031 & 5: \textbf{p=0.0002} & 6: \textbf{p=0.0229} \\

\hline
\multicolumn{12}{l}{\footnotesize $\uparrow$: Higher values are better, $\downarrow$: lower values are better \quad\quad\quad Color scale (normalized for each metric): \printThreeColorSpectrum} \\
\multicolumn{12}{l}{\footnotesize \textsuperscript{a} Van der Laan items reported as Mean (M) $\pm$ SD. All other measures reported as Median (Mdn) and IQR.}\\
\multicolumn{12}{l}{\footnotesize \textsuperscript{b} 1: HH-HR, 2: HH-RH, 3: HH-RR: 4: HR-RH, 5: HR-RR, 6: RH-RR}\\
\multicolumn{12}{l}{\footnotesize $\circ$ Minor violation of normality (Shapiro-Wilk test). No action taken since RM-ANOVA is robust against minor violations due to the Central Limit Theorem.} \\
\multicolumn{12}{l}{\footnotesize $\circ\circ$ Non-parametric test used (Friedman) following major violation of normality (Shapiro-Wilk test).} \\
\multicolumn{12}{l}{\footnotesize $\dagger$ Greenhouse-Geisser correction applied following violation of sphericity (Mauchly's test).} \\
\end{tabular}%
}
\label{tab:statistical_analysis}
\vspace{-4mm}
\end{table*}


\section{Discussion}
This study investigated the relevance of Fitts' classical ``Men Are Better At / Machines Are Better At'' (MABA-MABA) framework~\cite{fitts_human_1951} in the context of two-agent pHRC. The central question aimed to identify the optimal static allocation of position and force control between a human and a robot during physical collaboration. The findings presented here provide compelling evidence suggesting that the fundamental principles articulated by Fitts over seven decades ago retain significant relevance for designing effective collaboration strategies in modern pHRC scenarios involving direct physical interaction and shared control. Beyond this validation, the results also yield additional insights into user experience.

\begin{figure}[!t]
    \centering
    \includegraphics[trim={0.5cm 0.5cm 0.3cm 0.45cm},clip,width=0.90\linewidth]{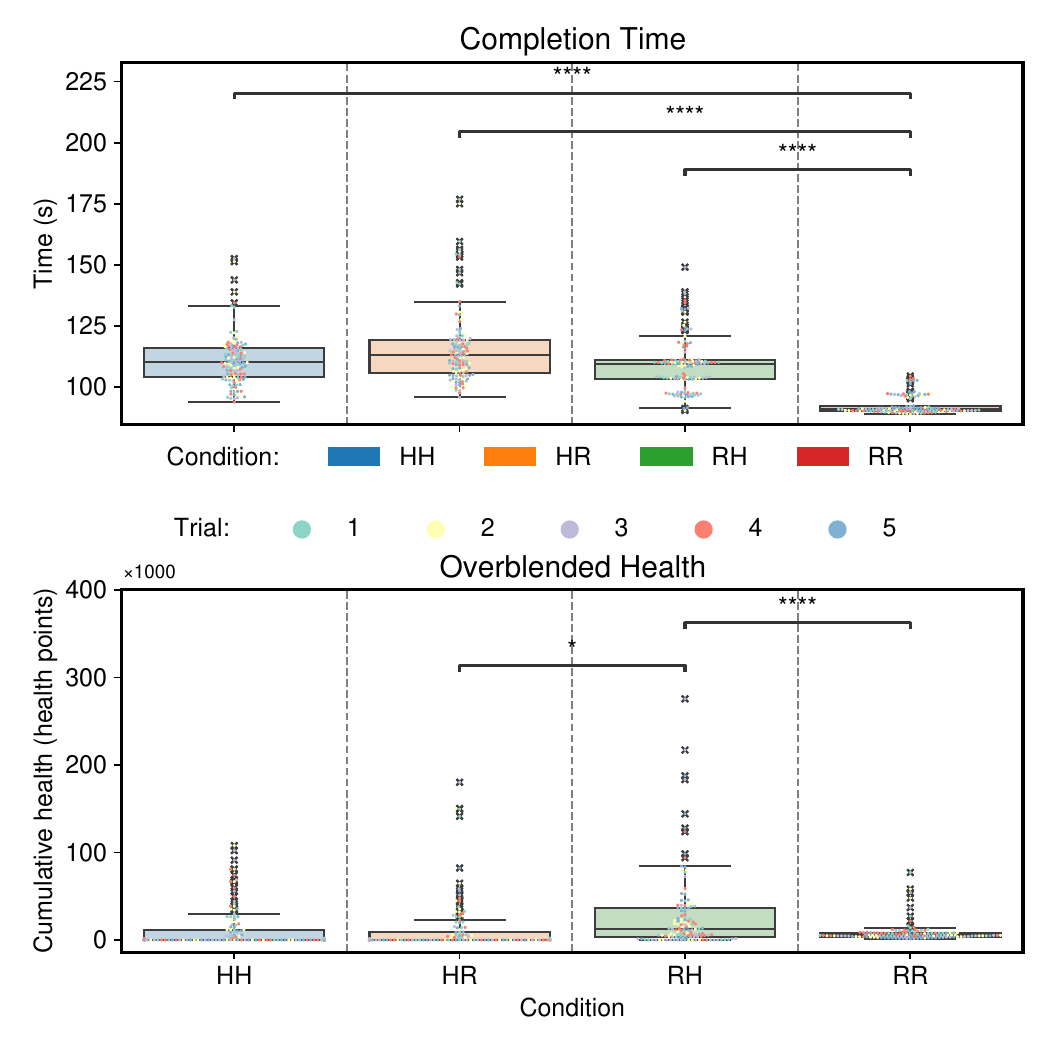}
    \caption{Box‑plots showing objective performance metrics in terms of completion time in seconds (top) and cumulative overblended health in health points (bottom). Boxes show the median and the interquartile range, where whiskers extend to the 5th and 95th percentiles, while each small dot is one trial and is colored according to the trial number. Lower values correspond to better performance. Statistically significant pair‑wise differences are annotated, where: *:~$0.01 < p \le 0.05$; ****:~$p \le 0.0001$.}
    \label{fig:performance}
    \vspace{-4mm}
\end{figure}

\begin{figure}[!t]
    \centering
    \includegraphics[trim={0.5cm 0.5cm 0.45cm 0.45cm},clip,width=0.90\linewidth]{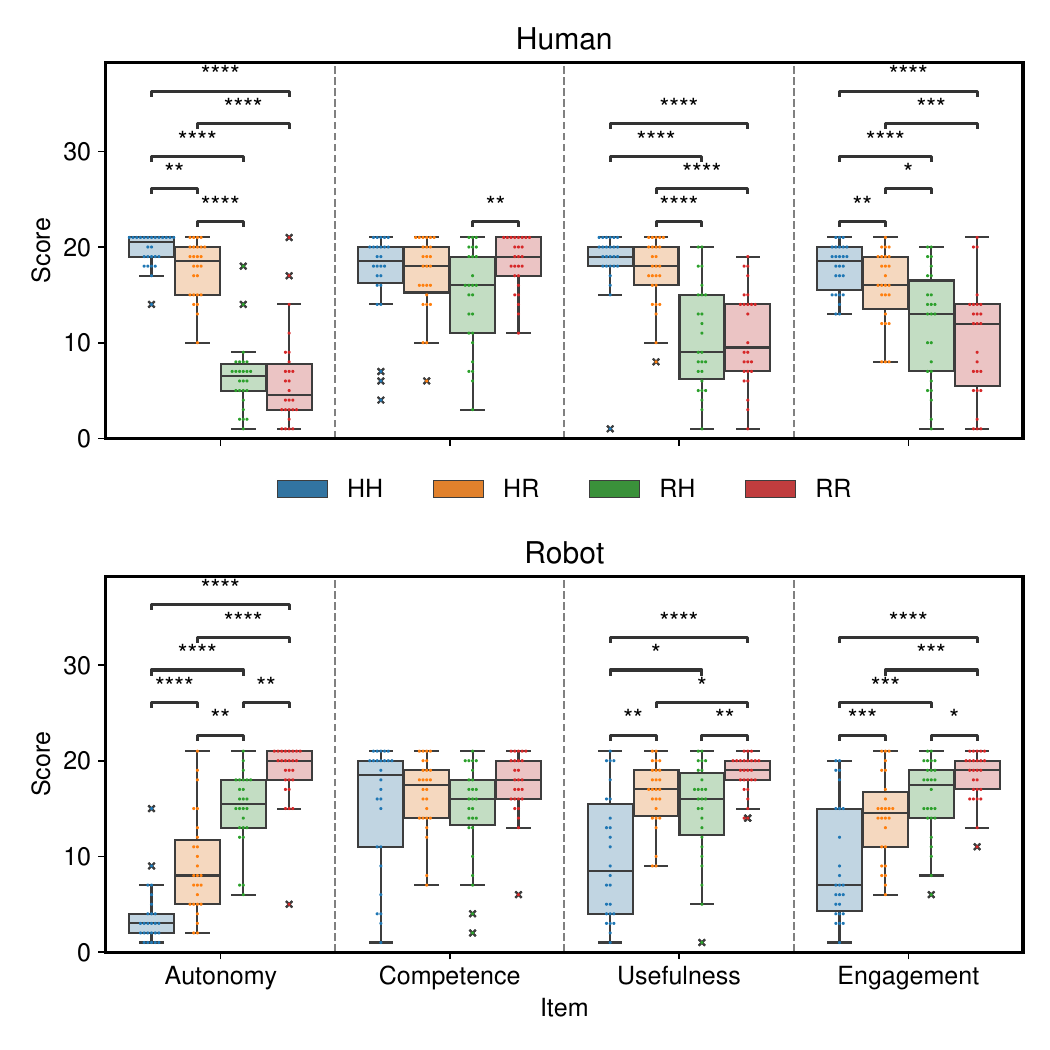}
    \caption{Box‑plots showing how participants rated themselves (top) and how they rated the robot (bottom) on four dimensions: autonomy, competence, engagement, and usefulness. For every condition, boxes show the median and the interquartile range, where whiskers mark the 5th–95th percentiles, while small dots show individual ratings. Statistically significant pair‑wise differences are annotated, where: *:~$0.01 < p \le 0.05$; **:~$0.001 < p \le 0.01$; ***:~$0.0001 < p \le 0.001$; ****:~$p \le 0.0001$.}
    \label{fig:mol-cis}
    \vspace{-4mm}
\end{figure}

\textbf{Validating Fitts. HR outperforms RH}:
The direct comparison between the HR and RH conditions consistently favored HR across multiple dimensions. Participants reported higher acceptance (supporting H1, Fig.~\ref{fig:vdl}), and demonstrated significantly less overblending, while completion time was found to be similar (largely supporting H2, Fig.~\ref{fig:performance}). This equivalence in completion time likely reflects a balance of two different inefficiencies: the time lost to the human's imperfect exploratory path in HR was matched by the time the robot in RH spent on corrective re-blending due to the human's imperfect force application for this specific task.
Participants also experienced significantly less frustration and lower physical demand (largely supporting H3, Fig.~\ref{fig:nasa-tlx}), and experienced greater autonomy, felt more useful, and engaged (strongly supporting H4, Fig.~\ref{fig:mol-cis}) when they controlled positioning while the robot controlled the force. These findings align well with the ``MABA-MABA'' principle: leveraging human dexterity and adaptability for position control and delegating force to the machine leads to a more effective and preferred collaboration.

\textbf{RH leads to increased overblending and frustration}:
The difficulties encountered in the RH condition, notably increased overblending and elevated frustration, suggest a misalignment. This could stem from the robot dictating the position trajectory, potentially introducing unpredictability, or a path that conflicts with the human's natural movement tendencies or intentions. This most likely forces the human operator to be reactive, attempting to synchronize their force output with a robot-imposed motion profile. A dip in self-rated competence (although not found to be statistically significant, see Fig.~\ref{fig:mol-cis}) could result from a loss in confidence due to this misalignment.

\textbf{HR provides better experience compared to HH}:
Interestingly, the HR condition was not only preferred over the inverse task allocation (RH), but was also rated as significantly more useful and satisfying than the HH baseline (Fig.~\ref{fig:vdl}), suggesting the Fitts-aligned collaboration (HR) provides a genuine enhancement over unassisted work. This is likely due to the substantial reduction in physical demand (Fig.~\ref{fig:nasa-tlx}), without compromising the user's sense of agency over the task's positional execution or a large drop in perceived autonomy (see HH vs HR and RH in Fig.~\ref{fig:mol-cis}).

Conversely, when comparing RH to HH, significantly lower satisfaction ratings were possibly due to a significant increase in frustration. Usefulness was rated similarly low, perhaps due to perceived higher physical demand. The participants rated themselves significantly less useful in the collaboration and felt less engaged. The latter aspect is interesting since one would expect that a human having to be more reactive to a robot would increase the engagement.

It is important to note that neither HR nor RH improved objective task performance compared to HH (Fig.~\ref{fig:performance}). The absence of a performance increase could imply that robot assistance is superfluous. Nevertheless, pHRC has many other benefits that do not directly affect the performance. For example, pHRC improves ergonomics in physically demanding tasks, such as blending, to prevent work-related musculoskeletal injuries~\cite{peternel_robot_2018}.

\textbf{Supervisory control and the ironies of automation}:
The RR condition presented a unique profile characterized by trade-offs. Objectively, RR showed the best completion times and the lowest cumulative overblended health, as it was designed to do (Fig.~\ref{fig:performance}). It also resulted in the lowest perceived workload across almost all NASA-TLX dimensions (Fig.~\ref{fig:nasa-tlx}). However, these efficiency gains came at a cost: participants reported the lowest levels of perceived autonomy, usefulness, and engagement (Fig.~\ref{fig:mol-cis}). This resonates with Bainbridge's classic ``ironies of automation''~\cite{bainbridge_ironies_1983}: optimizing system performance can paradoxically diminish human experience by reducing meaningful involvement.

Nevertheless, our findings, specific to pHRC, also partially contradict that classical notion since RR achieved surprisingly favorable acceptance ratings, ranking second overall, statistically comparable to HH, though still lower than HR (Fig.~\ref{fig:vdl}). One explanation is the substantial reduction in physical and temporal demands was highly valued by participants. Moreover, the supervisory role, requiring vigilance and occasional intervention, likely provided more engagement (perceived similarly to RH, see Fig.~\ref{fig:mol-cis}) than purely passive oversight, mitigating the worst effects of automation. Participants also rated the robot in RR as more useful and engaging than in HH and RH (Fig.~\ref{fig:mol-cis}), suggesting they recognized its effectiveness even while acknowledging their own reduced role.

Finally, this study assessed short-term task performance. In industrial settings, prolonged supervision may shift the trade-off between efficiency and meaningful involvement, potentially reducing acceptance over time.

\textbf{The primacy of position control for experienced autonomy}:
Results reveal a pronounced asymmetry in how delegation affected participants' sense of autonomy (Fig.~\ref{fig:mol-cis}). Delegating position control to the robot (HH vs RH) resulted in a substantially larger decrease in reported autonomy compared to relinquishing force control (HH vs HR), suggesting that position control is more intrinsically linked to the operator’s sense of agency than force control.

Position control determines \textit{where} the interaction occurs and directs the tool’s action towards achieving the task goal, whereas force control \textit{modulates} the intensity of the interaction at a given location. Participants may perceive the goal-directed aspect as the primary locus of control in executing the physical task.
Relinquishing positional authority may thus feel akin to losing command over task progression, downgrading the human contribution to one of reactive effort modulation.

This drop in autonomy when delegating position control aligns well with Self-Determination Theory (SDT) in a pHRC context. SDT posits that autonomy is a fundamental psychological need, and its frustration thwarts motivation and well-being, which can lead to negative outcomes such as burnout. Our findings empirically mirror this principle: RH recorded the lowest levels of perceived autonomy (shared with RR) and also yielded the highest levels of frustration and the lowest satisfaction scores. This suggests that for participants, delegating position control was not just a change in task mechanics but the removal of a psychological need, which resulted in a negative experience.

\textbf{Introduction of robot does not affect mental demand}:
The non-significant difference in mental demand suggests a comparable cognitive load, albeit of a different nature: HH required focused motor control for both sub-tasks, HR required coordination with the robot producing force, RH demanded a more reactive vigilance to synchronize with the robot's movement, and RR necessitated supervisory vigilance. This suggests that introducing a robot in any function does not make the task largely more mentally demanding, which can be a valuable insight for workplaces considering pHRC solutions.

\textbf{Limitations}:
Several limitations warrant consideration. The predominantly male participant pool (4 female, 22 male) restricts generalisability, and the abstracted lab task may not fully transfer to complex industrial settings where optimal task allocation could differ. Furthermore, our evaluation was based on short-term interaction; long-term use, especially in the supervisory (RR) condition, could alter user experience due to factors like sustained vigilance. Finally, as the new ACES questionnaire lacks full psychometric validation, its findings should be interpreted with caution.

\section*{Acknowledgment}
The authors would like to thank Alexis Derumigny for advice on the statistical evaluation.

\bibliographystyle{IEEEtran}
\bibliography{bibliography}

\end{document}